\documentclass{article}
\usepackage[T1]{fontenc} % add special characters (e.g., umlaute)
\usepackage[utf8]{inputenc} % set utf-8 as default input encoding
\usepackage{ismir,amsmath,cite,url}
\usepackage{graphicx}
\usepackage{amsmath}
\usepackage{amssymb}
\usepackage{color}

\usepackage{lineno}
\title{An Empirical Evaluation of End-to-end Polyphonic Optical Music Recognition}
\multauthor
{Sachinda Edirisooriya \hspace{1em} Hao-Wen Dong \hspace{1em} Julian McAuley \hspace{1em} Taylor Berg-Kirkpatrick}
{University of California San Diego\\
\small\texttt{\{\href{mailto:sediriso@ucsd.edu}{sediriso},\href{mailto:hwdong@ucsd.edu}{hwdong},\href{mailto:jmcauley@ucsd.edu}{jmcauley},\href{mailto:tberg@ucsd.edu}{tberg}\}@ucsd.edu}
}

\def\authorname{Sachinda Edirisooriya, Hao-Wen Dong, Julian McAuley and Taylor Berg-Kirkpatrick}

\usepackage[bookmarks=false,pdfauthor={\authorname},pdfsubject={\papersubject},hidelinks]{hyperref}
\usepackage{booktabs}
\usepackage[capitalize,noabbrev]{cleveref}

\begin{document}

\maketitle

\begin{abstract}
Previous work has shown that neural architectures are able to perform optical music recognition (OMR) on monophonic and homophonic music with high accuracy. However, piano and orchestral scores frequently exhibit polyphonic passages, which add a second dimension to the task. Monophonic and homophonic music can be described as homorhythmic, or having a single musical rhythm. Polyphonic music, on the other hand, can be seen as having multiple rhythmic sequences, or voices, concurrently. We first introduce a workflow for creating large-scale polyphonic datasets suitable for end-to-end recognition from sheet music publicly available on the MuseScore forum. We then propose two novel formulations for end-to-end polyphonic OMR---one treating the problem as a type of multi-task binary classification, and the other treating it as multi-sequence detection. Building upon the encoder-decoder architecture and an image encoder proposed in past work on end-to-end OMR, we propose two novel decoder models---FlagDecoder and RNNDecoder---that correspond to the two formulations. Finally, we compare the empirical performance of these end-to-end approaches to polyphonic OMR and observe a new state-of-the-art performance with our multi-sequence detection decoder, RNNDecoder.
\end{abstract}

\section{Introduction}\label{sec:introduction}

\begin{figure}
  \centering
  \includegraphics[width=\linewidth]{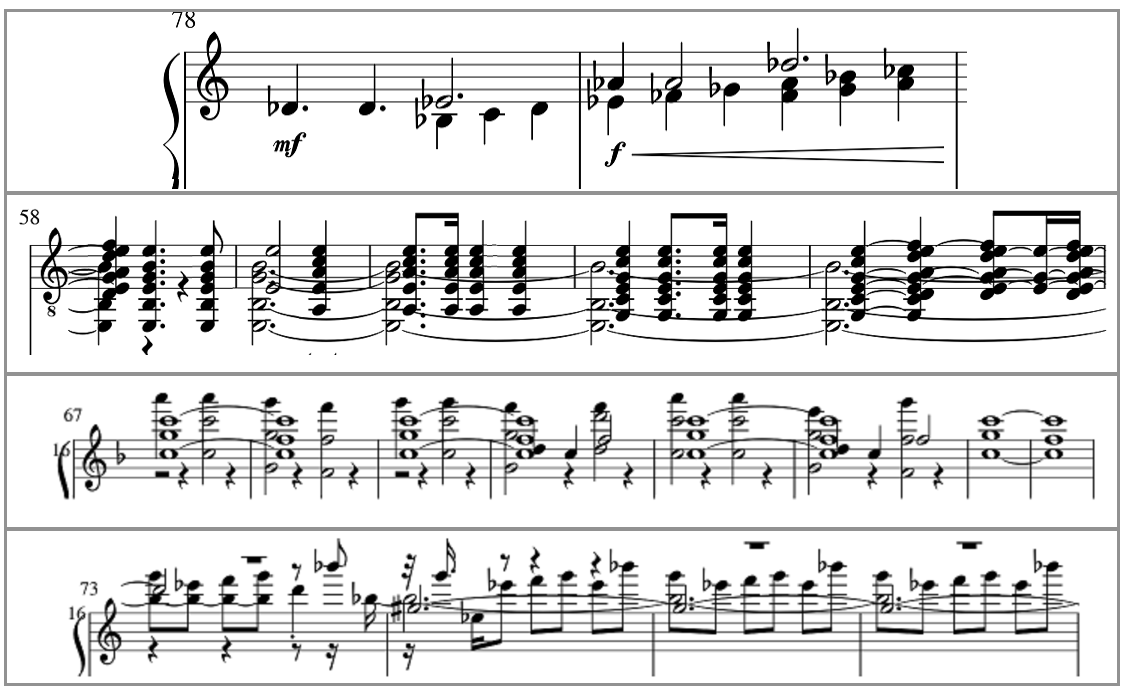}%\\
%   (1) An easy excerpt\\
%   (2-4) Some difficult excerpts
  \caption{Examples of the MuseScore Polyphonic Dataset (MSPD) and its hard subset (MSPD-Hard)---(top) an easy excerpt in MSPD and (bottom) three excerpts that can be found in both MSPD and MSPD-Hard.}
  \label{fig:examples}
\end{figure}

% Introduce why OMR is important
As society continues to become more and more dependent on digitization as a means of storing information such as photos, addresses, and music, now is a perfect time to refine the technology required to digitize sheet music. Organizing scanned music can be a tedious task to do manually. For example, each image must be labeled with several attributes, the most basic being the title, composer, arranger, and page number. Assuming that all of these are input correctly, a user could then navigate through a large-scale collection somewhat easily. However what if a user wanted to filter the scores by attributes such as instrument, key signature, time signature, or tempo? What if a user had a score, but wanted to transpose it to a different key?
%(a common practice in vocal-related performances)
Manually doing all of the annotation required for these demands when uploading sheet music scans would be impractical, and this is where optical music recognition (OMR) can shine.

% Mention current approaches
Over the past few years, data driven approaches to optical music recognition have become attractive ways to solve the problem. The improvement in the accuracy of systems built using these tools is very exciting, however they are far from perfect in challenging circumstances. One visual challenge relatively unique to optical music recognition is detecting multi-voice music, also known as polyphonic music. Previous work has mentioned that their approaches to OMR cannot sufficiently solve this problem to the same extent as they have solved monophonic OMR~\cite{calvo2020understanding,baro2019optical}.

% Introduce our work (MuseScore data)
In view of the lack of a large-scale polyphonic dataset for end-to-end polyphonic OMR, we introduce a workflow for acquiring annotated samples from sheet music publicly available on the MuseScore forum~\cite{musescore}. As an initial attempt to the challenges of polyphonic OMR, we only consider single-staff scores. 
Our objective with this dataset is to empirically evaluate the performance of three different architectures on the task of end-to-end polyphonic OMR, where the input is a staff line image and the output is a symbolic sequence that can be encoded into common music notation formats such as MusicXML. Voice information is not included in this process.

% Introduce our work (architectures)
We propose two novel neural architectures for end-to-end OMR, namely FlagDecoder and RNNDecoder, which are both encoder-decoder models based on the architecture proposed by Calvo-Zaragoza et al. for monophonic scores~\cite{calvo2017end}, hereafter referred to as the baseline. One feature common to all three models is their mechanism for encoding each image: a CNN followed by a bidirectional LSTM. The architectures differ, however, in their decoding mechanism. The baseline architecture uses a fully connected layer as a symbol classifier on the encoded image. Our proposed FlagDecoder treats polyphonic OMR as a multi-task binary prediction problem, simultaneously detecting whether each pitch and staff symbol is present or not, along with the rhythm if the symbol is a note. Our other proposed architecture, RNNDecoder, uses a vertical RNN decoder over the symbols appearing at each vertical image slice, giving the model the capacity to output a predetermined number of symbol predictions at a single horizontal position of an image.

% Briefly talk about our experiments
In this paper, we first introduce the current state of polyphonic OMR research.
% and the various approaches.
Then we introduce our procedure for building a large-scale dataset of exclusively polyphonic music data. Finally we perform an empirical evaluation of our proposed architectures introduced above, and find that they both outperform past work in terms of symbol error rate, with the RNNDecoder achieving a new state-of-the-art performance on end-to-end polyphonic OMR. All source code is available at \url{https://github.com/sachindae/polyphonic-omr}.

\section{Background}\label{sec:bg}

\subsection{Optical music recognition (OMR)}

Previous work on polyphonic OMR has been limited. One of the main approaches to it has been a two-step process of first segmenting each musical symbol~\cite{pacha2018handwritten,tuggener2018deep,hajic2018towards,pacha2018baseline,huang2019state} and then identifying the relationships between them through a post-processing step~\cite{pacha2019learning}. The challenge with building a system using this method is that errors add up from both sub-systems. Particularly the latter task, more formally known as forming a musical notation graph, can be quite challenging, with the state-of-the-art being far from perfect~\cite{pacha2019learning}. Another approach to OMR has been to treat it as an end-to-end task as proposed by~\cite{calvo2017end,vanderwel2017monomusescore}, where the complete symbolic sequence corresponding to an image is output by the system. Among the end-to-end approaches, there have been a variety of training objectives used for the task such as Connectionist Temporal Classification (CTC) loss, cross-entropy loss, and Smooth-L1 loss~\cite{calvo2017end,vanderwel2017monomusescore,baro2020handwritten}. While applied to monophonic and homophonic music~\cite{alfaro2019approaching} successfully, there have not been any conclusions on the framework's capability of being extended to polyphonic notation. 

\subsection{Datasets for OMR}

Several datasets have been proposed for musical symbol classification~\cite{desaedeleer2006openomr,fornes2008symbol,rebelo2010symbol,calvozaragoza2014homus,pacha2017universal,hajic2017muscima}. Others have been proposed for training end-to-end OMR systems on single-stave handwritten music scores~\cite{baro2019optical} and typeset images for monophonic scores~\cite{calvozaragoza2018primus,calvozaragoza2018cameraprimus,vanderwel2017monomusescore}. However, these datasets are either small in size or contain only monophonic scores.

\begin{table*}
  \centering
  \begin{tabular}{lllllll}
    \toprule
    &\multicolumn{3}{l}{MSPD}   &\multicolumn{3}{l}{MSPD-Hard}\\
    \cmidrule(lr){2-4} \cmidrule(lr){5-7}
                &Min &Mean &Max 
                &Min &Mean &Max\\
    \midrule
    Length (symbols)
        &3  &70.59 &819
        &41 &79.2  &679 \\ 
    Length (measures)
        &1  &4.15  &55
        &1  &2.11  &8\\
    Density (symbols/measures)
        &3  &20.3  &165
        &41 &52.8  &165 \\
    Polyphony (voices) 
        &2  &2.05  &4 
        &2  &2.42  &4\\
    \bottomrule
  \end{tabular}
  \caption{Statistics of the MuseScore Polyphonic Dataset (MSPD) and its hard subset (MSPD-Hard).}
  \label{tab:statistics}
\end{table*}

\section{MuseScore Polyphonic Dataset}\label{sec:dataset}

\begin{figure}
  \centering
  \includegraphics[width=\linewidth]{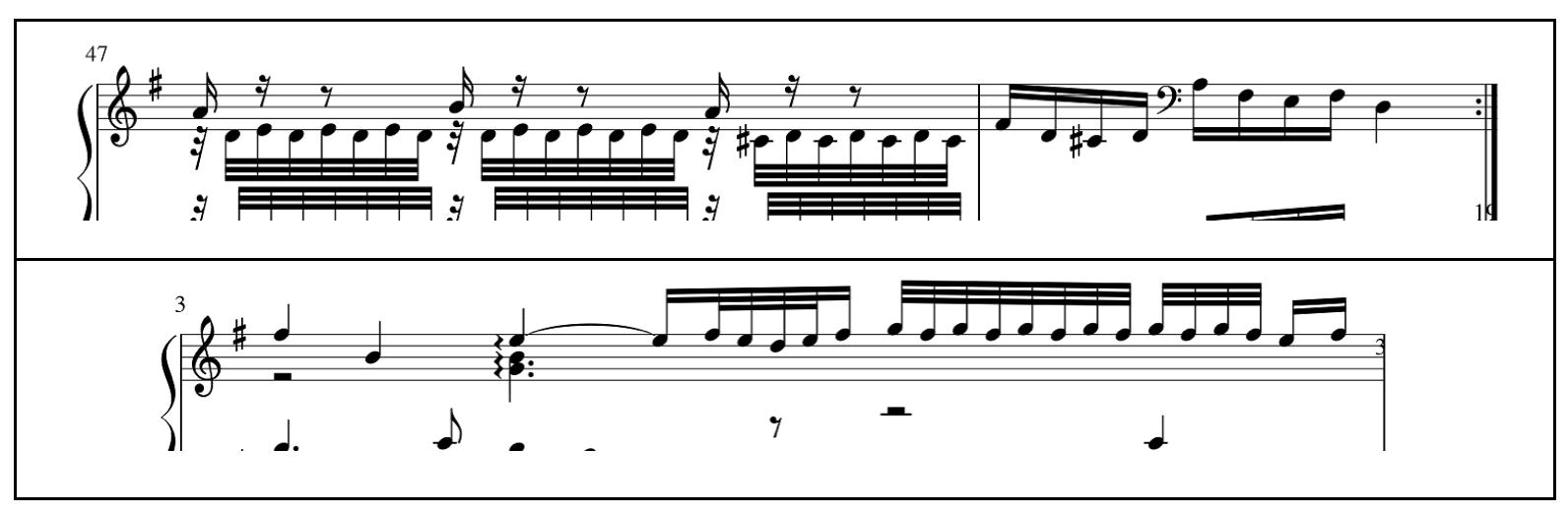}\\
  \caption{Examples from the dataset where the staff line is not cropped perfectly.}
  \label{fig:cropexample}
\end{figure}

\begin{figure}
  \centering
  \includegraphics[width=\linewidth]{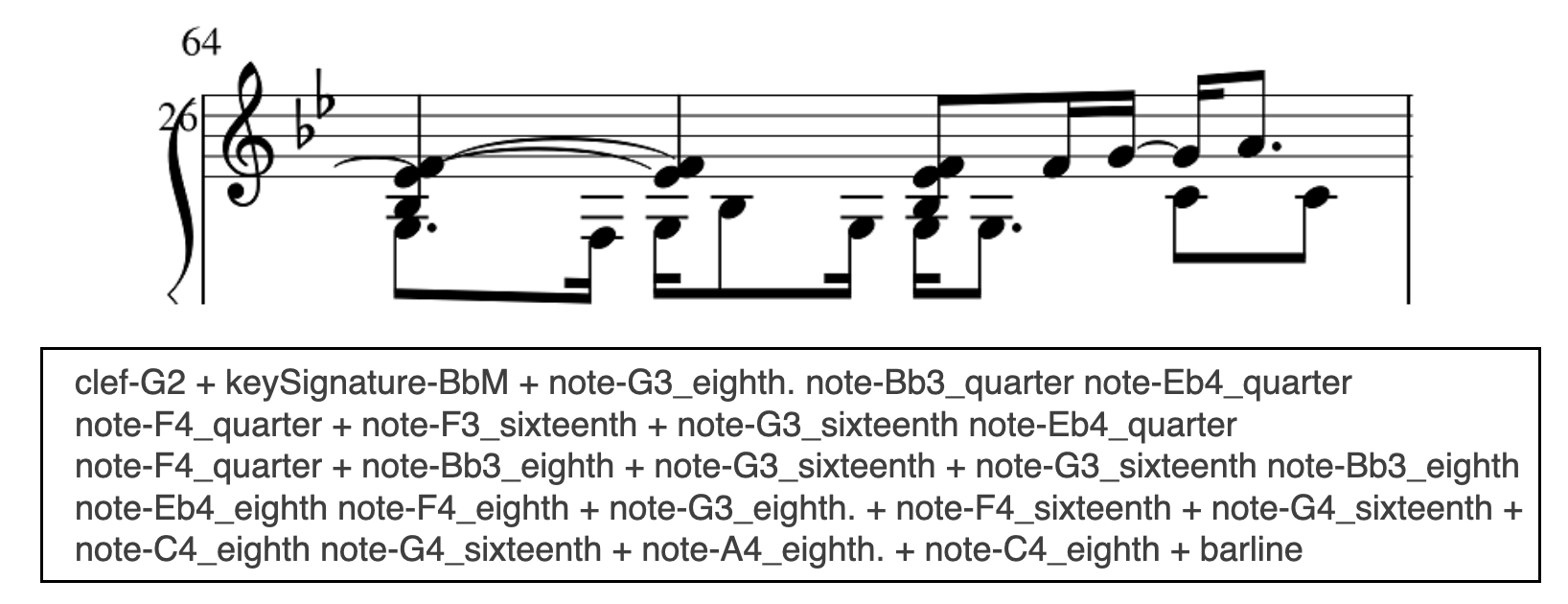}\\
  \caption{Example of the label encoding adopted in this paper.}
  \label{fig:gtexample}
\end{figure}

Given the size of datasets such as PrIMuS~\cite{calvozaragoza2018primus} that have been used to show the effectiveness of an end-to-end architecture on monophonic OMR, we decided to come up with a dataset of similar size to do the same for polyphonic OMR. Using 19,432 MuseScore files available to download online~\cite{musescoredataset}, we were able to generate 124,671 varied-width single-staff images of exclusively polyphonic music along with their ground truth labels. To determine whether a given sample is polyphonic or not, we check if the music encoding defines multiple voices within a single measure.

To generate the dataset, we first used the MuseScore software plugin API (separate from the MuseScore forum mentioned above) to reduce the page height of the rendering to easily generate single staff line images. Then we executed a script to remove credit text which covered up some music symbols. After that, we used the MuseScore plugin API to generate MusicXML and PNG files from the MuseScore files. Lastly, we parsed the MusicXML to generate labels for each of the images, and removed sparse (music with only rests) and non-polyphonic data from the dataset.

The difficulty of the samples has quite a large range, from simple excerpts with two-note chords to dense notation as shown in \cref{fig:examples}. Also, many samples are not perfectly cropped resulting in an additional implicit task of extracting the staff line from noisy environments. We argue however that this creates a more realistic environment as OMR systems should be able to handle these kinds of interferences shown in \cref{fig:cropexample}. 
 
For our experiments with this dataset, we used a 70/15/15 split for training, validation, and test respectively.

\subsection{Data annotation}\label{sec:data_labeling}

Due to our primary focus being polyphony, we chose to use a minimal symbol set sufficient to represent pitch and rhythm accurately, apart from tuplets. More specifically, we do not care about symbols such as dynamics, ties, tuplets, staccatos, accents, and other staff text. Instead, the only musical symbols we chose to label are clefs, key signatures, time signatures, barlines, and the notes themselves (pitch and rhythm). Inspired by the labeling scheme used to train models for end-to-end monophonic and homophonic OMR~\cite{calvo2017end,alfaro2019approaching}, we approached the task of polyphonic OMR in the same way, with the ``Advance position'' encoding proposed by Alfaro et al. for representing homophonic music as a one-dimensional sequence reading from left to right. This encoding adds a `+' symbol between each sequential occurrence of notes and symbols, and orders the individual notes of a chord from bottom to top, as seen in \cref{fig:gtexample}. Throughout the rest of this paper, we will refer to the non-note music symbols described above (clefs, key signatures, time signatures, and barlines) as staff symbols.

\subsection{MSPD-Hard}

While we are interested in the performance on the average polyphonic images, we also want to have a means to push an OMR system to its limit so we can better determine an upper bound on the capabilities. To do this, we created a subset of the test set of all the samples with a density (defined by number of symbols per measure) of at least 41, resulting in 900 samples, which we will refer to as MSPD-Hard. The bottom three images in \cref{fig:examples} are samples from the hard subset.

The statistic that sticks out the most when comparing the two (see \cref{tab:statistics}) is of course the density. While it may seem strange that the mean symbol length of the hard test set is similar to the full test set, we observed that this is because the more dense samples tend to contain measures that are filled with symbols, thus fewer measures can fit per image on average. %It also may seem alarming that there is a sample with 55 measures in it, however this is due to the varied width of the images in the dataset. 
Lastly, the hard subset samples tend to have a higher level of polyphony, measured by number of voices defined by the MusicXML files, as expected.

\begin{figure*}
  \centering
  \includegraphics[width=\linewidth]{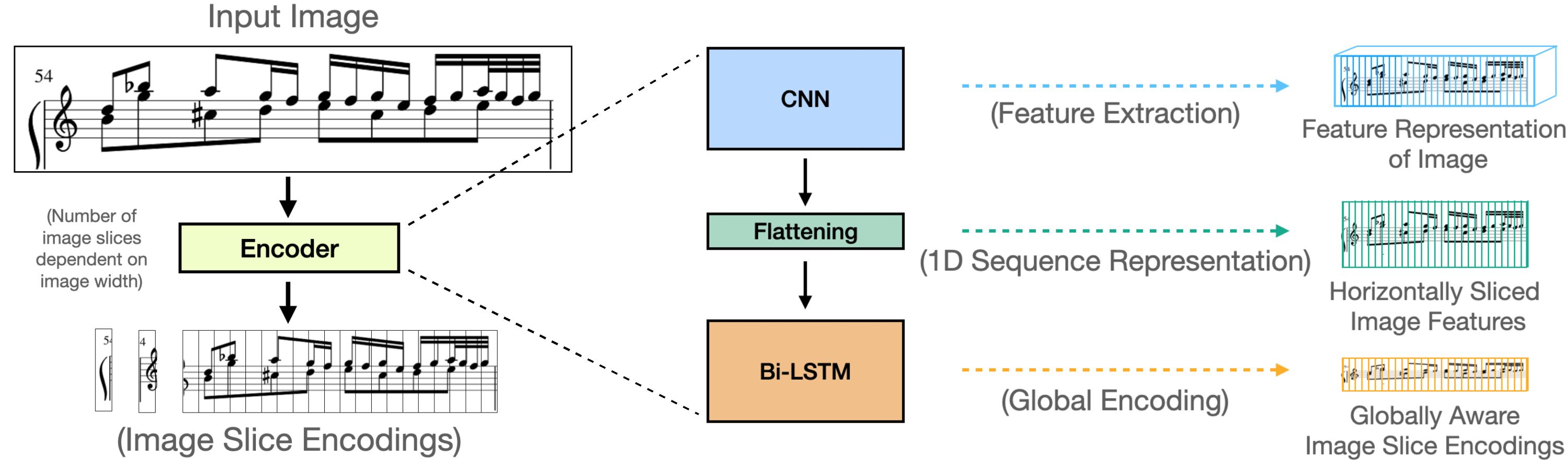}
  \caption{Illustration of the image encoder used in this paper.}
  \label{fig:encoder}
\end{figure*}

\section{Architectural Comparison for Polyphonic Optical Music Recognition}

As we mentioned in \cref{sec:bg}, several training objectives have been used for end-to-end OMR. Due to many recent works showing the effectiveness of CTC~\cite{graves2006connectionist} for OMR~\cite{calvo2017end,baro2019optical,alfaro2019approaching}, we chose to train all of the considered architectures using the same objective. The equation below shows the objective that CTC aims to maximize, where $y$ represents the target sequence, $z$ the alignment, $x$ the sequence of vertical image slices (i.e. fixed-width slices of a staff line as shown in \cref{fig:encoder}), and $\theta$ the model parameters.

\begin{equation}
    \max_{\theta}P(y\,|\,x; \theta) = \max_{\theta}\sum_{z} P(y,z\,|\,x; \theta)\,.
\end{equation}

\subsection{Architecture overview}

At a high level, all of the architectures we compared share the same structure. The two components are an encoder and a decoder. The function of the encoder is firstly to extract features about the image while creating vertical slices through pooling, and secondly to give global context of the image encoding to each local image slice. The goal of the decoder is to use the representations created by the encoder and predict the symbols that are present in each of the vertical image slices. Finally, we use CTC to marginalize over the alignment of these slices for training. Due to the effectiveness of this end-to-end architecture on monophonic and homophonic music ~\cite{calvo2017end,alfaro2019approaching}, we chose to keep the encoder fixed and instead look to different decoding strategies that could be better suited for dense polyphonic music.

\subsection{Encoder details}\label{encoder_sec}

The encoder we use is the one described by Calvo-Zaragoza et al, with 2x less width pooling~\cite{calvozaragoza2018cameraprimus}. First, the image is fed through a deep convolutional neural network (CNN) to extract relevant learned features. These two-dimensional feature representations of each vertical image slice are then flattened to a vector
% single dimension
to create a sequential representation suitable for a recurrent neural network (RNN). The purpose of the RNN in this context is to give some awareness of the surrounding representations to each image slice's encoding, which is essential to help identify symbols that may span multiple image slices. We chose to use a bidirectional Long Short-Term Memory (LSTM)~\cite{hochreiter1997lstm,schuster1997bilstm} for its effectiveness in end-to-end OMR as empirically shown by Baro et al.~\cite{baro2019optical}. A visualization of the encoder components can be seen in \cref{fig:encoder}.

\subsection{Baseline decoder}

The baseline architecture we evaluated includes the encoder mentioned above, followed by a simple decoder consisting of two parallel fully connected layers to classify the pitch and rhythm of the symbol appearing in each vertical image slice. By design, this decoder is only capable of outputting a single symbol prediction at each image slice, a characteristic not ideal for polyphonic music. 

\subsection{FlagDecoder}
% Intuition
% To model the multi-dimensional nature of polyphonic music while being trained on a one-dimensional sequence, we propose the BinaryVector. We observed that there are a fixed number of positions on a staff line that a note can appear on, indicating its pitch when combined with knowledge of the clef, key signature, and any accidentals. Based on this, one possible decoding for a vertical image slice could be an $n$-dimensional (where $n$ is the sum of the number of note positions and the number of staff symbols) vector where the value of each dimension indicates whether or not the corresponding pitch or staff symbol appears in the image slice. This can be achieved using the sigmoid activation function. When performing inference, we apply a threshold of 0.5 to determine if a symbol is present or not.

% However since in OMR, it is not enough to just identify the note positions on a staff line that are present, but also their rhythms and pitch modifiers such as accidentals, we modify this BinaryVector approach so that each note position on the staff has a corresponding rhythm and accidental classifier as opposed to a binary classifier that is only applicable for staff symbols. Visualizing this in \cref{fig:decoder}, there is a binary vector for the staff symbols, and a matrix for the notes, together resembling a flag hence the name FlagDecoder. To handle the case when two voices are playing the same note, we include two rows in the note matrix for each note position on the staff.

To incorporate the multi-dimensional nature of polyphonic music while being trained on a one-dimensional sequence, we propose the BinaryVector. We observed that there are a fixed number of positions on a staff line that a note can appear on, indicating its pitch when combined with knowledge of the clef, key signature, and any accidentals. Based on this, one possible decoding for a vertical image slice could be an $n$-dimensional (where $n$ is the sum of the number of note positions and the number of staff symbols) vector where the value of each dimension indicates whether or not the corresponding pitch or staff symbol appears in the image slice. This can be achieved using the sigmoid activation function. For inference, we apply a threshold of 0.5 to determine if a symbol is present or not.

\begin{figure*}
  \centering
  \includegraphics[width=.8\linewidth]{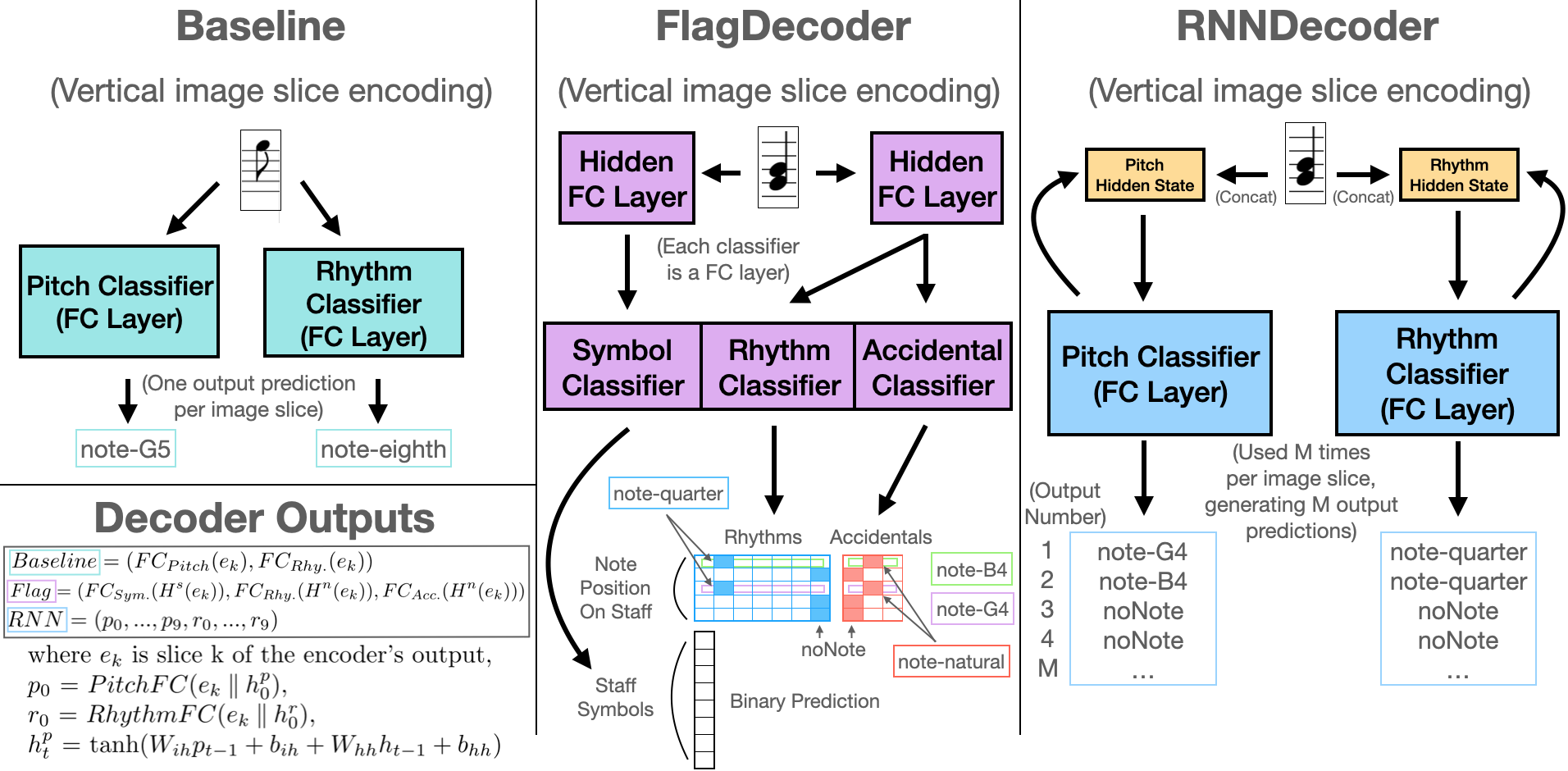}\\
  \caption{Illustrations of the three decoder architectures examined in this paper.}
  \label{fig:decoder}
\end{figure*}

Since we also need rhythm and accidental information for OMR, we modify this BinaryVector approach so that each note position on the staff has a corresponding rhythm and accidental classifier as opposed to a binary classifier that is only applicable for staff symbols. Visualizing this in \cref{fig:decoder}, there is a binary vector for the staff symbols, and a matrix for the notes, together resembling a flag hence the name FlagDecoder. To handle the case when two voices are playing the same note, we include two rows in the note matrix for each note position on the staff.

% Implementation
The FlagDecoder first uses two parallel fully connected (FC) layers for the note positions and staff symbols to reduce the dimensionality. Then the note latent representation goes through two separate FC layers (for rhythms and accidentals) to produce the note matrix, and the staff symbol latent representation is connected to another FC layer to produce the BinaryVector. %The main difference with this decoder's implementation from the baseline implementation is that when optimizing with CTC, each ``symbol" does not correspond to a single word from the vocabulary -- rather each possible flag (notes and staff symbols combination) is its own ``symbol".

%\begin{equation}
%    P(s_j | x_i) = P(r_j | x_i) * P(a_j | x_i) * P (w_j | x_i)
%\end{equation}
%where $r_j$ represents rhythm predictions, $a_j$ represents accidental %predictions, and $w_j$ represents staff symbol predictions.

\subsection{RNNDecoder}
\begin{figure}
  \centering
  \includegraphics[width=\linewidth]{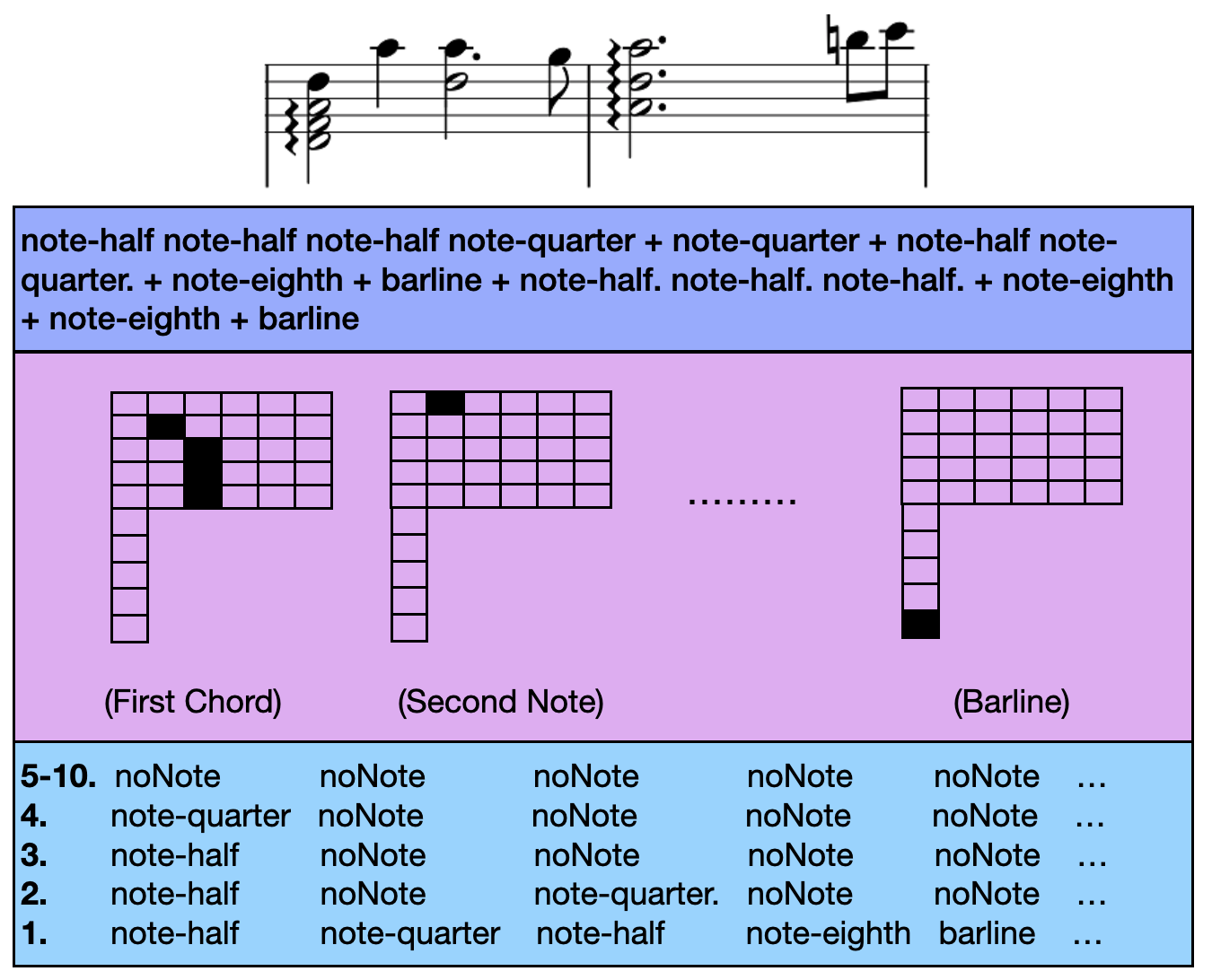}\\
  \caption{Examples of the different symbolic sequence representations used to optimize the decoders---(top) BaselineDecoder using ``Advance position'' encoding, (middle) FlagDecoder and (bottom) RNNDecoder, where number of outputs is 10.}
  \label{fig:encodings}
\end{figure}

% Intuition explanation
An alternative approach to those mentioned above, which maintain a single dimensional sequence, is to embrace the fact that polyphonic music is inherently multi-dimensional. Based on the near perfect results that have been shown in previous published work on monophonic OMR with the baseline decoder, we took a new approach to the challenge of polyphony that breaks up the task into several simpler versions. Rather than representing polyphonic music as a one-dimensional sequence, we propose RNNDecoder, a recurrent decoder that is run across each image slice vertically, allowing for multiple outputs at a single image slice. 
This can ideally handle polyphony better than the baseline decoder, 
% This approach views polyphonic music as consisting of multiple one-dimensional sequences, 
and can be trained trivially using the perspective that there are multiple one-dimensional sequences occurring from bottom to top in each image. To allow for alignment at inference time, we add a ``noNote'' symbol to the labels, as shown in \cref{fig:encodings}.

% Implementation explanation 
%The RNNDecoder implementation is similar to the baseline decoder in the sense that it has a single classifier each for pitch and rhythm classification. 
As we mentioned above, 
%this decoder
the RNNDecoder can generate a fixed number $m$ of outputs per image slice. Since the largest number of notes and staff symbols we observed present in a single horizontal position of an image was 10, we chose to have $m = 10$ for our experiments. The key difference between the RNNDecoder and the baseline is that a hidden state is included while generating each output prediction, and is concatenated to the current image slice encoding before going through the classifier. More specifically, at each image slice as each output is generated, this hidden state is updated like the hidden state of an RNN as in %\cref{eqn:multiseq}
\cref{fig:decoder}, and fed through the same classifier along with the image slice encoding 10 times. As shown in \cref{fig:encodings}, this new decoder can be trained without using the ``Advance position'' label encoding. %For optimization, we minimize the arithmetic mean of the CTC losses over the $m$ sequences.
%\begin{equation}\label{eqn:multiseq}
%    h_t = \tanh(W_{ih}x_t + b_{ih} + W_{hh}h_{t-1} + b_{hh})\,.
%\end{equation}
%where $h_t$ is the hidden state at step t, $x_t$ is the output of the classifier at step t, and $W$ and $b$ correspond to learned weight matrices and biases respectively.

\subsection{Objective functions}

% Basic CTC explanation
As we mentioned in Section 4, CTC aims to maximize the likelihood of a target symbolic sequence. For the baseline decoder, we use two target sequences -- one for rhythm and one for pitch -- and jointly minimize those two CTC losses. On the other hand for the FlagDecoder, we just use a single target sequence where each "symbol" is a unique flag configuration (see \cref{fig:encodings}), meaning it represents a combination of notes and staff symbols as opposed to just a single word from a symbol vocabulary. Lastly for the RNNDecoder, we use the same two target sequences used in the baseline, but we use 10 of them, thus optimizing the arithmetic mean of the CTC losses over the $m$ sequences. The loss functions are shown below in \cref{eqn:loss}:
\begin{equation}\label{eqn:loss}
\begin{aligned}
\mathcal{L}_{Baseline} &= \mathcal{L}_{Pitch} + \mathcal{L}_{Rhythm}\,,\\
\mathcal{L}_{FlagDecoder} &= \mathcal{L}_{Flag}\,, \\
\mathcal{L}_{RNNDecoder} &= \frac{1}{m}\sum_i^m{\left(\mathcal{L}_{Pitch} + \mathcal{L}_{Rhythm}\right)}\,, \\
\end{aligned}
\end{equation}
where $\mathcal{L}_{Pitch}$, $\mathcal{L}_{Rhythm}$ and $\mathcal{L}_{Flag}$ are the corresponding CTC losses.

\section{Experiments}

\begin{table*}
  \centering
  \begin{tabular}{lllll}
    \toprule
    &\multicolumn{2}{l}{MSPD}   &\multicolumn{2}{l}{MSPD-Hard}\\
    \cmidrule(lr){2-3} \cmidrule(lr){4-5}
    Decoder model    &Rhythm SER (\%) &Pitch SER (\%) &Rhythm SER (\%) &Pitch SER (\%)\\
    \midrule
    Baseline         &7.39            &10.28           &14.11           &18.83\\
    FlagDecoder      &6.67            &9.82           &9.86           &17.98\\
    RNNDecoder       &\textbf{3.92}            &\textbf{5.64}           &\textbf{5.82}            &\textbf{8.57}\\
    \bottomrule
  \end{tabular}
  \caption{Performance of the three decoders on MuseScore Polyphonic Dataset (MSPD) and its hard subset (MSPD-Hard).}
  \label{tab:result}
\end{table*}

\subsection{Implementation details}\label{subsec:training}

All models were trained using stochastic gradient descent with the Adam optimization algorithm~\cite{kingma2014adam}. We used a learning rate of $10^{-4}$ and a batch size of 16. Additionally, each image goes through a short preprocessing stage where they are inverted so that black pixels take on the highest value while white pixels are 0, and subsequently resized to a fixed height of 128 pixels.

\subsection{Experiment setup}

For evaluation, we follow recent literature~\cite{baro2019optical} and use Pitch and Rhythm Symbol Error Rate (SER) as our evaluation metric of choice. This metric measures the edit distance at a symbol level between a predicted sequence and the ground truth sequence. More precisely, it can be written as the sum of insertions ($I$), deletions ($D$), and substitutions ($S$) normalized by the ground truth sequence length ($N$), i.e., $\text{SER} = (I + D + S)/N$.
% \begin{equation}
%     SER = \frac{I + D + S}{N}
% \end{equation}

In addition to staying consistent with previous literature, we also chose to evaluate Pitch and Rhythm SER separately to be able to compare the difficulties of the two tasks and gain insight on where improvements can be made. The two sets of data we used for evaluation are the full test set containing 18,700 images, and the hard test set containing 900 images, both of which were discussed in depth in \cref{sec:dataset}.

\subsection{Experiment results}

We first compare the three different decoders using the full test set. An interesting result is that the Pitch SER is higher than the Rhythm SER across the board for all of the models. For the baseline and RNNDecoder, we believe this is due to the fact that pitch is affected by clef, thus when a clef prediction is off, or in particularly wide images, the correct clef may not be represented in an image slice encoding, resulting in incorrect decoding. Additionally when there are multiple neighboring notes together, identifying the correct pitches naturally seems more challenging than identifying the correct rhythms due to requiring more fine grained features to discern the exact pitches. We believe using an agnostic representation of pitch as described in~\cite{calvozaragoza2018primus} could potentially result in better pitch performance. 

\begin{figure}
  \centering
  \includegraphics[width=\linewidth]{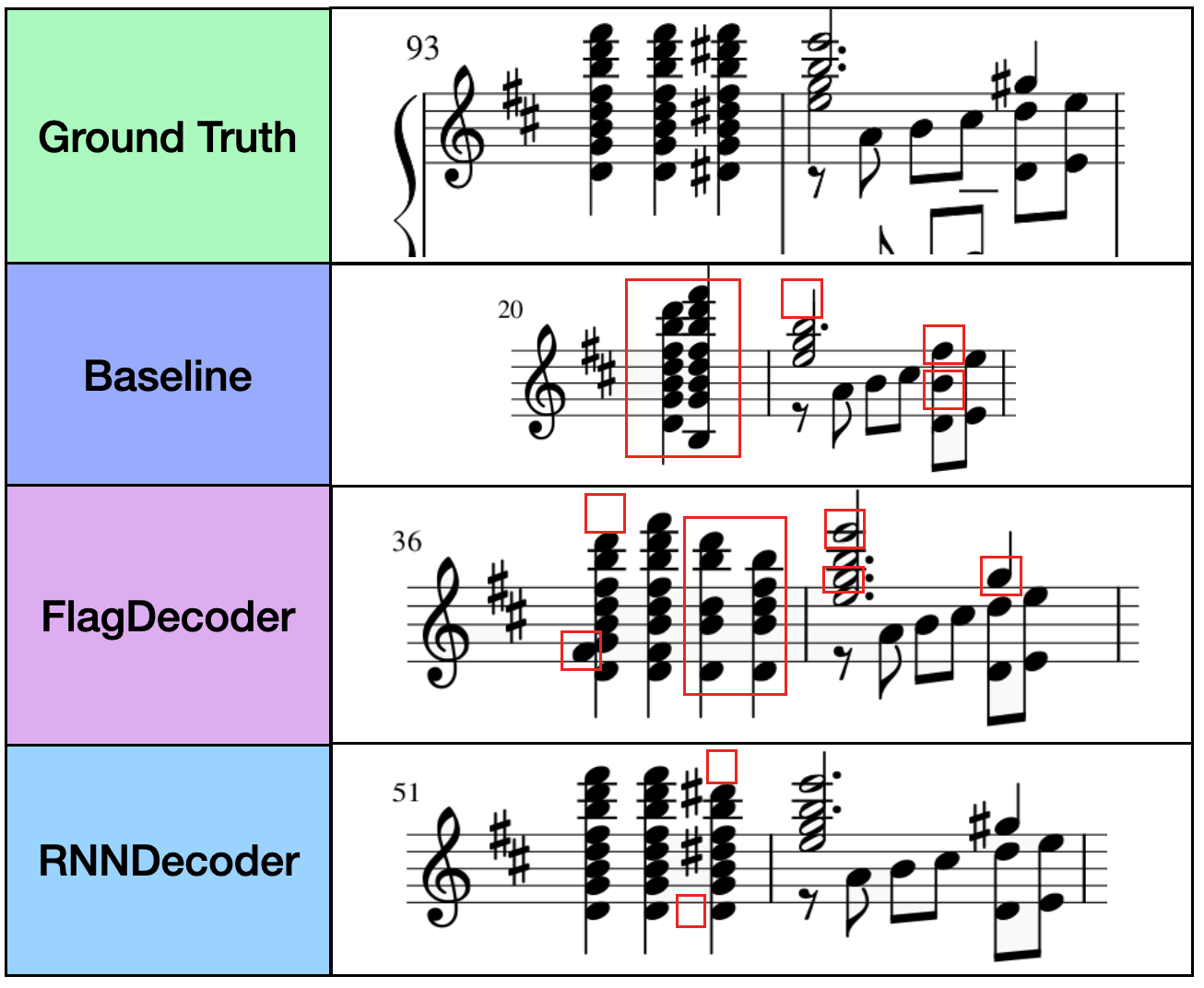}
  \caption{Comparison of the three decoders on an example with dense chords. Red blocks indicate the errors.}
  \label{fig:qual1}
\end{figure}

While our FlagDecoder only slightly outperforms the baseline on the full test set, the RNNDecoder appears to perform twice as well as the baseline on the full test set, achieving a new state-of-the-art performance. Further evaluating the performances of the new decoding methods on the hard test set (MSPD-Hard) highlights our decoders' strengths over the baseline decoder. With the baseline, the error rate nearly doubles, whereas with the proposed decoding strategies, the error rate increases by a smaller factor when dealing with difficult data, with the Pitch SER of the FlagDecoder being the exception.

\subsection{Error analysis}

\begin{figure}
  \centering
  \includegraphics[width=\linewidth]{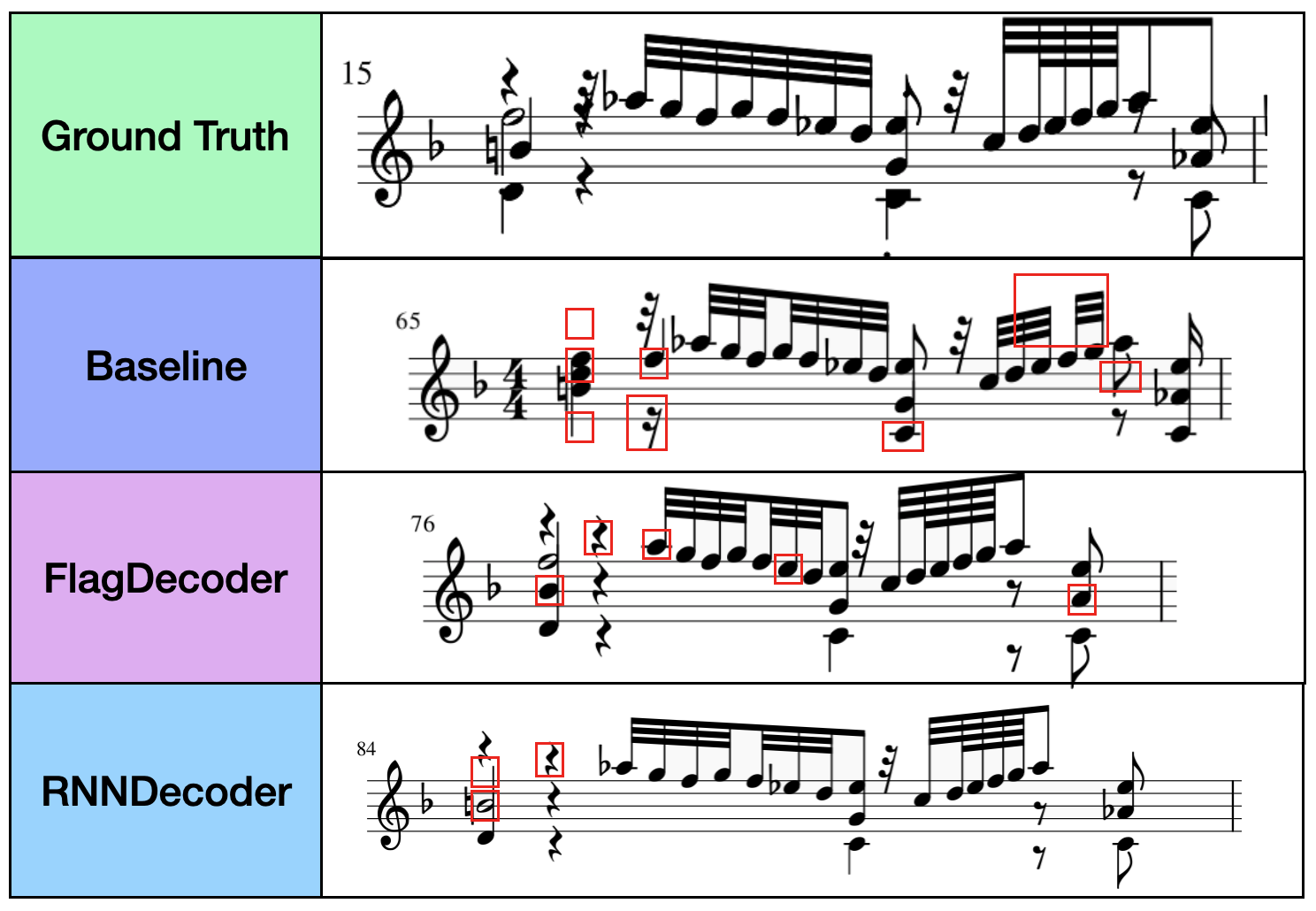}
  \caption{Comparison of the decoders on an example with four levels of polyphony. Red blocks indicate the errors.}
  \label{fig:qual2}
\end{figure}

We also examine some qualitative results on examples from MSPD-Hard to examine the upper bound performance of the decoders. From \cref{fig:qual1} we see that in the first measure which contains several huge chords nearby, the baseline is neither able to separate them by outputting a `+' symbol nor output the correct number of notes. The FlagDecoder handles the first two chords well, but is likely thrown off by the sharps in the third chord. The RNNDecoder performs the best in the first measure, just missing a single note and accidental in the last chord, and deals with the polyphony in following measure perfectly. \cref{fig:qual2} is challenging due to its high level of polyphony, which the baseline decoder clearly struggles with. In this example, both the FlagDecoder and RNNDecoder perform similarly, however the FlagDecoder actually handles the first chord with 4 voices correctly rhythmically whereas the RNNDecoder makes a minor mistake.

\section{Conclusion}

% Summary of contributions
In this work, we first introduced a workflow for generating datasets from MuseScore files which will be beneficial for the research community given the massive amount of
sheet music publicly available on the MuseScore forum.
% MuseScore files available.
We then used this to create a large-scale dataset suitable for end-to-end polyphonic
optical music recognition (OMR), and proposed two novel decoding strategies for the task, namely FlagDecoder and RNNDecoder. Further we performed an empirical comparison of the performance of these methods on polyphonic OMR, and observed a new state-of-the-art performance with the RNNDecoder.

% Limitations
One of the main limitations of this work is generalizability, a problem that many supervised machine learning systems struggle with. While our system can perform extremely well on images generated from the MuseScore engraver, it is not able to do well out of the box on music generated by other engravers.
% Future work
We are also aware that there are many viable implementations of the new decoding methods we proposed that could potentially give better performance, and hope to evaluate them in the future when we have the required computing resources. In addition, adapting these methods to multi-staff music will allow more versatility in usage. Lastly, we hope to address the challenge of generalizability in the future as it is currently one of the major barriers preventing neural network-based OMR systems from being deployed widely.

\bibliography{ref}

\end{document}